\pdfoutput=1

\documentclass[11pt]{article}

\usepackage{acl}

\usepackage{times}
\usepackage{latexsym}
\usepackage{amssymb}
\usepackage{amsmath}
\usepackage{booktabs}
\usepackage{graphicx}
\usepackage{subcaption}
\usepackage{xcolor}
\usepackage{float}

\usepackage[T1]{fontenc}

\usepackage[utf8]{inputenc}

\usepackage{microtype}

\title{Multi-Query Focused Disaster Summarization via \\Instruction-Based Prompting}

\author{Philipp Seeberger \and Korbinian Riedhammer \\
  Technische Hochschule Nürnberg Georg Simon Ohm \\
  \texttt{\{philipp.seeberger,korbinian.riedhammer\}@th-nuernberg.de}}

\begin{document}

\maketitle
\begin{abstract}
Automatic summarization of mass-emergency events plays a critical role in disaster management.
The second edition of CrisisFACTS aims to advance disaster summarization based on multi-stream fact-finding with a focus on web sources such as Twitter, Reddit, Facebook, and Webnews.
Here, participants are asked to develop systems that can extract key facts from several disaster-related events, which ultimately serve as a summary.
This paper describes our method to tackle this challenging task.
We follow previous work and propose to use a combination of retrieval, reranking, and an embarrassingly simple instruction-following summarization.
The two-stage retrieval pipeline relies on BM25 and MonoT5, while the summarizer module is based on the open-source Large Language Model (LLM) LLaMA-13b.
For summarization, we explore a Question Answering (QA)-motivated prompting approach and find the evidence useful for extracting query-relevant facts.
The automatic metrics and human evaluation show strong results but also highlight the gap between open-source and proprietary systems.
\end{abstract}

\begin{figure}[t]
    \centering
    \includegraphics[width=\columnwidth]{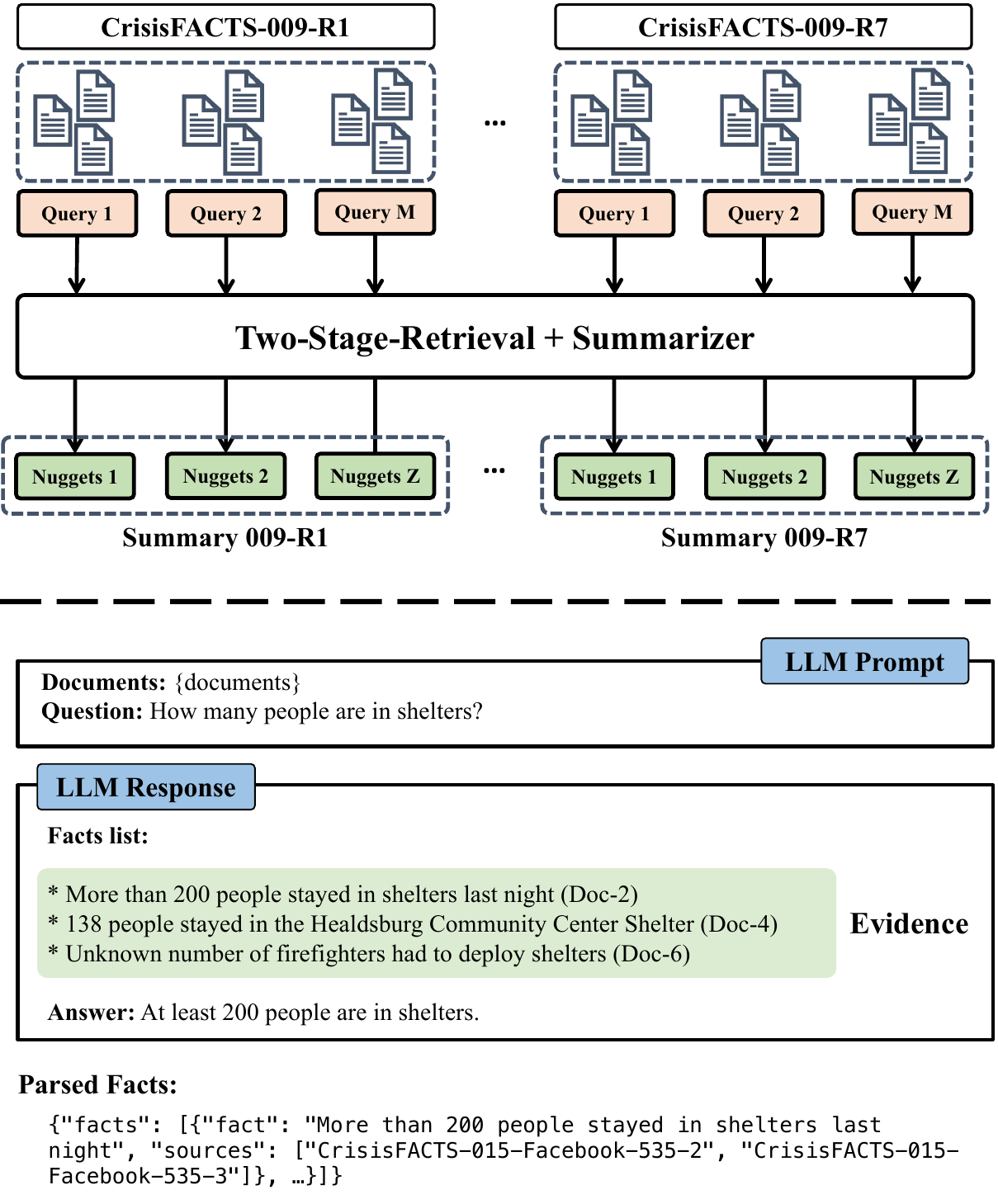}
    \caption{High-level overview of our proposed system and prompting strategy. The upper part depicts the overall pipeline. We call the pipeline for each event-request-query triple separately, resulting into the final summaries for each event-request pair. The lower part illustrates the prompting strategy. We generate query-focused facts using a QA-motivated approach and concatenate the extracted facts to form the final event nuggets.}
    \label{fig:overview}
\end{figure}

\section{Introduction}
Insufficient situational awareness during natural or human-made disasters can lead to significant loss of life, property, and environmental damage. 
Advancements in today's information ecosystem present new avenues for emergency response \cite{trecis_2021,kruspe_2021}.
For example, integrating heterogeneous online sources such as social media and microblogging platforms which rapidly disseminate crucial details about ongoing events \cite{sakaki_2010,reuter_2018}. 
This shift has created a multi-stream environment where traditional sources are augmented with emerging online platforms, recognized as a promising area in prior research efforts \citep{allan_1998,aslam_2015,sequiera_2018,trecis_2021}..

However, the rapid pace of content creation and the unique characteristics of various information sources and events pose challenges for existing models \cite{kaufhold_2021,seeberger_nlp4pi_2022}. 
The introduction of the CrisisFACTS Track aims to address these challenges, asking the community to develop summarization systems that can extract event nuggets for various events \cite{crisisfacts_2023}.

Recently, the use of instruction-following LLMs has attracted attention in various Natural Lanuage Processing (NLP) domains and zero-shot as well as in-context learning methods show competitive performance to traditional state-of-the-art models.
This is also evident in the disaster summarization domain, where extractive frameworks with decent performance \cite{seeberger_trec_2022} are surpassed by LLMs from proprietary APIs \cite{pereira_crisis_2023}.
We participate in the second edition of the CrisisFACTS challenge and propose to employ a simple but effective instruction-following approach, leveraging recent advances in open-source LLMs.

\paragraph{Results Summary}
Our proposed LLM-based event nugget generation approach achieves competitive performance and surpasses the majority of systems in the CrisisFACTS 2023 Track.
This trend is shown for both human and automatic evaluation results, underscoring the potentials of LLM-based disaster summarization.
However, our qualitative analysis reveals shortcomings in the generated query-related facts, suggesting the need for further development efforts.

\section{Model Description}\label{sec:method}
\figurename~\ref{fig:overview} gives an overview of the components of our system and how they are connected to generate the final summaries.
First, we retrieve and rerank the documents for each event-request-query triple to obtain query-relevant clusters of candidate documents.
Then, we use these clusters to extract query-relevant facts which serve as the basis for the event nuggets and final summaries.
We detail the components in the next sections.

\subsection{Retrieval and Reranking}
We follow a two-stage retrieval approach consisting of first-stage retrieval and subsequent reranking components.
This approach often has shown strong cross-domain performance abilities \cite{thakur2021beir}.
In the first step, we employ efficient lexical retrieval to reduce the computational costs for transformer-based models such as cross-encoders.
We first retrieve the top-$k^{(1)}$ candidate documents for each query represented as indicate terms.
Next, we process each query-related cluster of candidate documents with a neural reranker model.
Typically, these reranker models are cross-encoder variants.
Each query-related cluster is reranked by an additional query (e.g., question), and the top-$k^{(2)}$ are chosen to further reduce the set of candidate documents.
For further details, we refer to previous work \cite{seeberger_trec_2022}.

\subsection{\textsc{LLaMA-Nuggets}}
For summarization, we follow a QA-motivated approach and extract query-relevant facts (\figurename~\ref{fig:overview}).
In this way, we aim to filter out irrelevant documents and abstract only the query-relevant content from the document collections.
We prompt an instruction-following LLM for each event-request-query triple and use the retrieved and reranked candidate documents as well as corresponding query as input.
Here, the query represents the question, and we ask the model to provide a list of facts which serve as answer evidence.
This prompting scheme is also known as chain-of-thought (CoT).
We then parse the CoT fact items and cited documents into a structured format.
The referenced documents are important for two reasons: 1) Regarding traceability, the task requires providing the source documents of each event nugget. 2) We need an importance score that can be derived from the reranking relevance scores.
However, the resulting facts are rather short and atomic, and we aim to produce event nuggets covering a specific topic.

\paragraph{Event Nugget Generation}
To generate the event nuggets, we iteratively concatenate all generated facts for each specific query.
Here, we limit the character length to 200, which corresponds to the task's instructions.
As an importance score, we compute the mean of all relevance scores of the referenced documents.

\begin{table*}[t]
  \centering
  \resizebox{\linewidth}{!}{
  \begin{tabular}{lcccccccccc|c}
    \toprule
    $\downarrow$\hspace{0.2em}\textbf{Method}\hspace{1.5em}\textbf{Event}$\rightarrow$ & 009 & 010 & 011 & 012 & 013 & 014 & 015 & 016 & 017 & 018 & Avg
    \\\midrule
    \textsc{Greedy}$^{\dagger}$\hspace{0.8em} & 0.70 & 5.50 & 0.83 & 3.01 & 7.78 & 0.15 & 0.41 & 4.36 & 4.54 & 5.31  & 3.26
    \\
    \textsc{ILP-MMR}$^{\dagger}$\hspace{0.8em} & 0.52 & 5.37 & 0.80 & 2.37 & 6.62 & 0.40 & 1.54 & 4.10 & 5.14 & 5.12 & 3.20
    \\
    
    \textsc{LLaMA-Nuggets}\hspace{0.8em} & \textbf{15.77} & \textbf{9.56} & \textbf{10.77} & \textbf{20.32} & \textbf{17.42} & \textbf{14.57} & \textbf{15.84} & \textbf{18.90} & \textbf{18.46} & \textbf{18.28} & \textbf{15.99}
    \\\midrule\midrule

    \textsc{TREC-Mean}\hspace{0.8em} & 7.85\hspace{2pt} & 8.10\hspace{2pt} & 8.11\hspace{2pt} & 7.11\hspace{2pt} & 8.28\hspace{2pt} & 6.20\hspace{2pt} & 7.16\hspace{2pt} & 8.41\hspace{2pt} & 8.13\hspace{2pt} & 8.98\hspace{2pt} & 7.83\hspace{2pt}
    \\\bottomrule
  \end{tabular}}
  \caption{Comprehensiveness (i.e., recall) scores (x100) for our submitted system runs. $\dagger$ denotes that the corresponding system run was evaluated with automatic nugget matching. Bold numbers indicate the best performance.}
  \label{tab:human_recall}
\end{table*}

\begin{table*}[t]
  \centering
  \resizebox{\linewidth}{!}{
  \begin{tabular}{lcccccccccc|c}
    \toprule
    $\downarrow$\hspace{0.2em}\textbf{Method}\hspace{1.5em}\textbf{Event}$\rightarrow$ & 009 & 010 & 011 & 012 & 013 & 014 & 015 & 016 & 017 & 018 & Avg
    \\\midrule
    \textsc{Greedy}$^{\dagger}$\hspace{0.8em} & 14.29 & 13.69 & 1.54 & 22.44 & 46.49 & 7.14 & 4.76 & 20.88 & 30.09 & 47.37  & 20.87
    \\
    \textsc{ILP-MMR}$^{\dagger}$\hspace{0.8em} & 13.10 & 21.54 & 8.00 & 23.33 & \textbf{47.47} & 28.57 & 26.98 & 38.86 & 41.07 & 62.82 & 31.17
    \\
    
    \textsc{LLaMA-Nuggets}\hspace{0.8em} & \textbf{48.61} & \textbf{25.66} & 8.35 & \textbf{47.39} & 45.02 & \textbf{78.45} & \textbf{52.37} & \textbf{48.50} & \textbf{66.16} & \textbf{78.75} & \textbf{49.93}
    \\\midrule\midrule

    \textsc{TREC-Mean}\hspace{0.8em} & 25.08\hspace{2pt} & 18.08\hspace{2pt} & \textbf{10.11}\hspace{2pt} & 27.86\hspace{2pt} & 35.97\hspace{2pt} & 35.60\hspace{2pt} & 24.34\hspace{2pt} & 23.98\hspace{2pt} & 34.34\hspace{2pt} & 47.32\hspace{2pt} & 28.27\hspace{2pt}
    \\\bottomrule
  \end{tabular}}
  \caption{Redundancy (i.e., precision) scores (x100) for our submitted system runs. $\dagger$ denotes that the corresponding system run was evaluated with automatic nugget matching. Bold numbers indicate the best performance.}
  \label{tab:human_precision}
\end{table*}

\begin{figure*}[t]
    \centering
    \includegraphics[width=\textwidth]{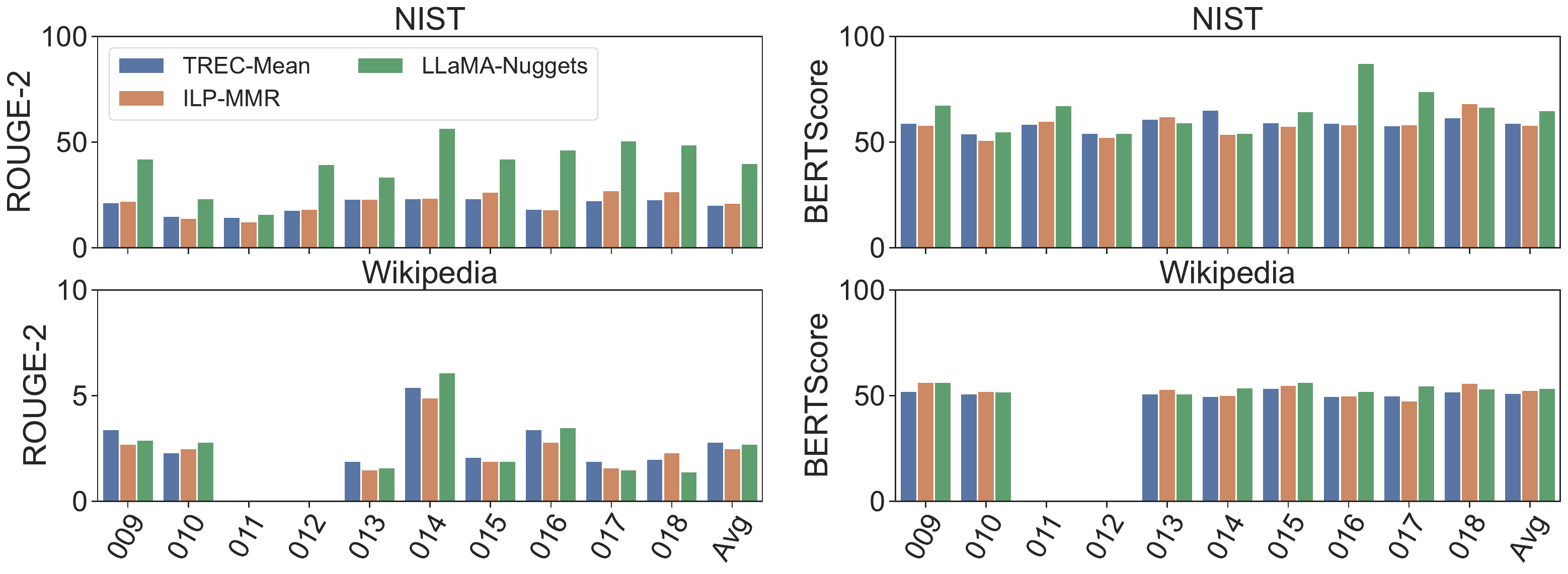}
    \caption{Rouge-2 and BERTScore $F1$-score (x100) results on reference summaries.}
    \label{fig:automatic}
\end{figure*}

\begin{figure}[t]
    \centering
    \includegraphics[width=\columnwidth]{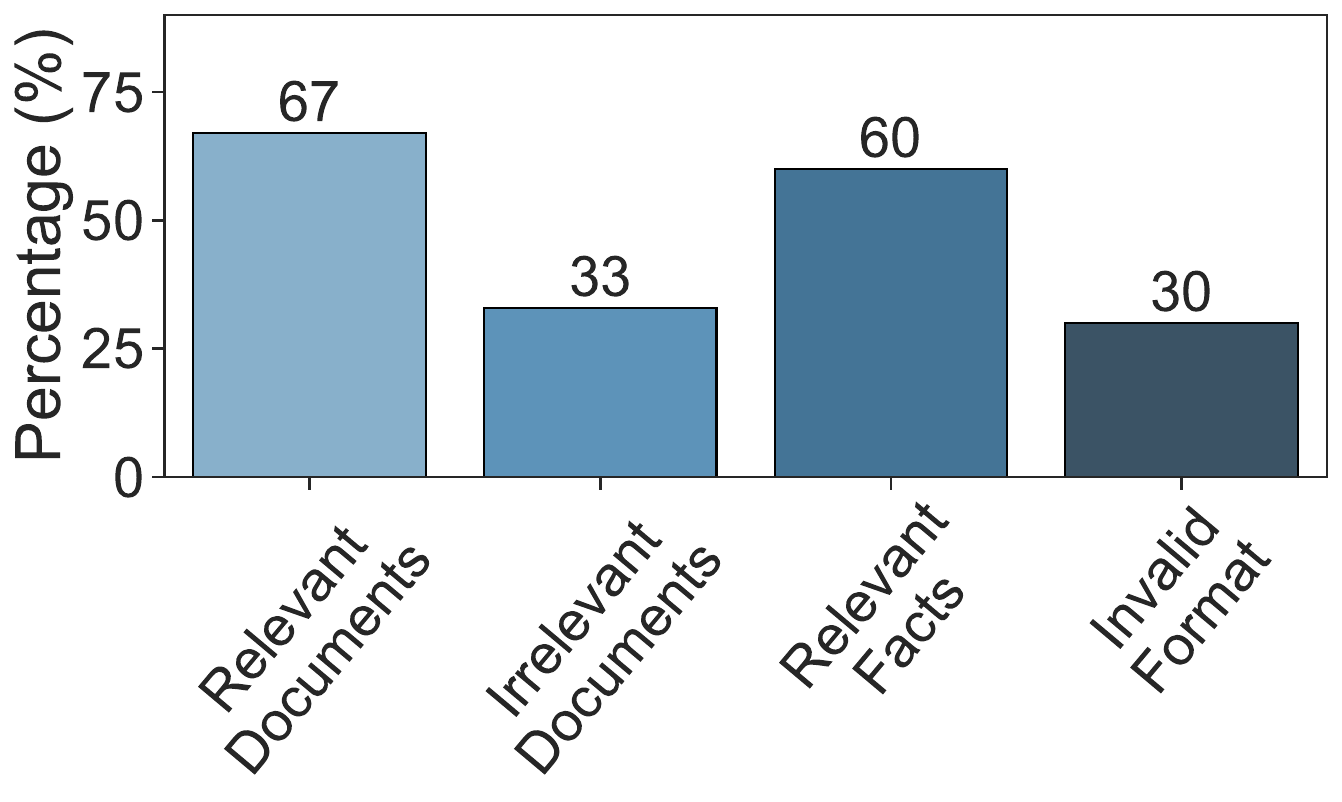}
    \caption{Qualitative analysis results for 30 LLM prompts and responses. \textbf{Prompts}: The fraction of prompts that contain at least one relevant or only irrelevant documents. \textbf{Responses}: The fraction of responses that contain at least one relevant generated fact or whether the output has an undesired format.}
    \label{fig:prompting_errors}
\end{figure}

\begin{figure}[t]
    \centering
    \includegraphics[width=\columnwidth]{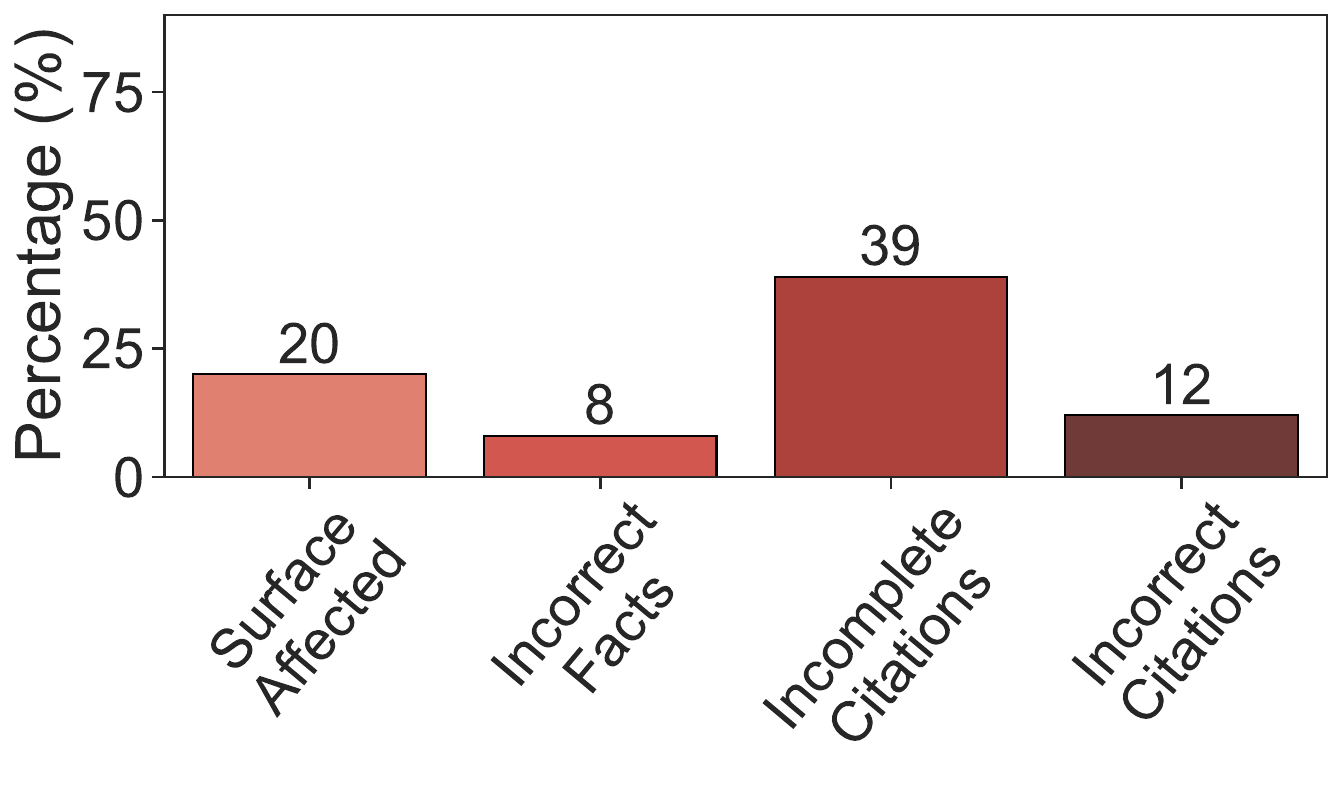}
    \caption{Qualitative analysis results for 59 generated facts. We show the fraction for surface issues, incorrect facts, and incomplete or incorrect citations.}
    \label{fig:facts_errors}
\end{figure}

\section{Experiments}

In the following, we detail the experimental setup including preprocessing, modeling, and evaluation.
Throughout our experiments, we consider all web sources provided by the TREC CrisisFACTS Track and reuse the already chunked stream items.

\subsection{Dataset}
The CrisisFACTS 2023 challenge includes a total of 10 disaster events and additional metadata (e.g., search keywords, queries, source types, etc.).
Each event is composed of multiple requests (i.e., days) covering multi-stream data extracted from online sources such as Twitter, Reddit, Facebook, and Webnews.
For further dataset details, we refer to the official CrisisFACTS website\footnote{\url{https://crisisfacts.github.io}}.

\subsection{Preprocessing}
For preprocessing, we normalize all Twitter posts to represent the text content similarly to other present online sources.
Specifically, we remove all retweet indicating prefixes, user mentions, emoticons, emojis, and URLs. 
Furthermore, we eliminate all hashtag symbols and split the text into corresponding words using the \textit{WordSegment}\footnote{\url{https://grantjenks.com/docs/wordsegment}} toolkit.
Lastly, we also remove exact duplicates. 
The majority of these duplicates are found in the tweet documents, mainly associated with retweets.

\subsection{Model Details}

As already mentioned in Section~\ref{sec:method} and shown in \figurename~\ref{fig:overview}, our pipeline is composed of the two-stage-retrieval and subsequent summarization modules.
Next, we describe the model and implementation details for each of these components.

\paragraph{\textsc{Retriever}} 
For first-stage retrieval, we follow previous work \cite{seeberger_trec_2022} and utilize the BM25 model with default settings from the \textit{PyTerrier} \cite{macdonald_2020} library.
To increase the recall, we extend it with Bo1 \cite{amati_2002} query expansion
and set the number of feedback terms and documents as 3 and 20, respectively. For each query, we concatenate the stemmed query text and indicative terms, and retrieve the top-$k^{(1)}=250$ candidate documents.

\paragraph{\textsc{Reranker}}
For second-stage reranking, we employ the MonoT5\footnote{castorini/monot5-large-msmarco-10k} model that is fine-tuned on the MS MARCO passage dataset.
MonoT5 is based on a pre-trained sequence-to-sequence model that generates relevance labels as target tokens \cite{nogueira_2020}.
In preliminary experiments, we find that this family of rerankers outperformed QA-motivated and encoder-based models \cite{seeberger_trec_2022}.
To reduce computational costs, we select the top-$k^{(2)}=30$ documents for the \textsc{LLaMA-Nuggets} and $50$ documents for the baseline models, respectively.

\paragraph{\textsc{LLaMA-Nuggets}}
For event nugget generation, we use the LLaMa-2-13B\footnote{meta-llama/Llama-2-13b-chat-hf} model series \cite{touvron2023llama} as underlying LLM, while relying on the fine-tuned version trained on multiple instruction datasets.
We also experimented with foundation models, but observed that these models fail to provide the desired output format.
Furthermore, we did not experience improvements with larger quantized versions such as 33B but found a drop in performance for the 7B parameters model.
For all experiments, we use the \textit{Transformers} \cite{wolf_2020} library, employ 4-bit quantization with normalized floats, and provide one demonstration sample.
The full prompt is shown in Appendix \ref{appendix:prompts}.

\paragraph{Baselines} 
In addition to the introduced \textsc{LLaMA-Nuggets}, we consider heuristic and extractive baselines.
\textsc{Greedy} simply selects the top-$k$ documents ordered by importance scores \cite{crisisfacts_2023}.
As extractive model, we include the last year's model \textsc{ILP-MMR} and similarly employ entities\footnote{\url{https://stanfordnlp.github.io/stanza/ner.html}} as concepts, frequency as weights, and set $L=150$ for the ILP formulation \cite{seeberger_trec_2022}.
For MMR, we set the trade-off parameter $\lambda=0.8$ and use TF-IDF for cosine similarity.

\subsection{Evaluation Metrics}
The CrisisFACTS organizers provide the evaluation results covering both automatic and human evaluation metrics.
Wikipedia excerpts and NIST summaries (constructed based on as useful annotated meta-facts) serve as gold standard summaries for automatic evaluation.
Here, system summaries represent the top-$k=32$ event nuggets for each event-request pair, which are evaluated with ROUGE-2 and BERTScore $F1$-scores.
Regarding human evaluation, the top-$k=20$ event nuggets are evaluated in terms of comprehensiveness and redundancy, as detailed in Appendix \ref{appendix:metrics}.

\section{Results}

\paragraph{Human Evaluation}
For human evaluation, we show the comprehensiveness and redundancy results in \tablename~\ref{tab:human_recall} and \tablename~\ref{tab:human_precision}, respectively.
Due to submission count limits for human evaluation, we conduct automatic event nugget matching for the \textsc{Greedy} and \textsc{ILP-MMR} baselines.
That is, we employ the BERTScore model to match the sytems' event nuggets with the meta-facts.
However, we observe that only a faction of event nuggets are matched and want to emphasize the unfair comparison\footnote{31.83\% for \textsc{Greedy} and 25.17 \% for \textsc{ILP-MMR}.}.
Nevertheless, we decided to include the results.
Our experimental results clearly demonstrate that \textsc{LLaMA-Nuggets} significantly outperform both the majority of TREC participants' systems as well as extractive baselines.
This trend is observed on 10/10 events for comprehensiveness and on 8/10 events for redundancy measures, while the comprehensiveness still indicates relatively low recall.
Overall, we present competitive results with a simple but effective LLM-based approach and demonstrate the potential of recent LLMs to improve the summarization of disaster events.

\paragraph{Automatic Evaluation}
In \figurename~\ref{fig:automatic}, we present the ROUGE-2 and BERTScore $F_1$-scores for NIST and Wikipedia event summaries.
Note that Wikipedia gold standard summaries are not available for events 011 and 012.
On average, \textsc{LLaMA-Nuggets} outperform the baselines \textsc{TREC-Mean} and \textsc{ILP-MMR} for all evaluation metrics except ROUGE-2 w.r.t. Wikipedia;
we hypothesize that it performs worse due to the nugget format, which is less fluent than, for example, Webnews extracts.
Interestingly, we observe that our model performs worse for event 014 in terms of BERTScore but achieve superior results for ROUGE-2.
These contrary results can be explained by entity surface form issues or the event nugget generation format.
However, the organizers used the BERTScore model DeBERTa\footnote{microsoft/deberta-xlarge-mnli}, which has a token limit of 512, while the event summaries exceed this limit by far.
Effectively, the BERTScore metric only evaluates the event nuggets of a subset of requests, which can lead to flawed evaluation results.

\paragraph{Qualitative Analysis} 
We randomly sample 30 event-request-query triple prompts (resulting in 59 generated facts) and qualitatively analyze both on the response and fact levels.
We find that 33\% of the prompt input documents did not contain any useful query-relevant information, highlighting the importance of noise robustness (\figurename~\ref{fig:prompting_errors}).
Only 60\% of the responses include at least one query-relevant fact, while 30\% show formatting issues.
In \figurename~\ref{fig:facts_errors}, we illustrate the errors at the fact level.
8\% are incorrect facts (i.e., hallucinations), 39\% miss relevant citations, and 12\% reference wrong documents.
We also check for redundancy issues related to entity surface forms and observe that 20\% of the assessed facts are affected.

\section{Conclusion}
In this work, we present our system for the TREC CrisisFACTS 2023 Track.
We combine two-stage retrieval consisting of an efficient sparse retriever and sequence-to-sequence reranker with instruction-following LLM summarization.
The experiments show that rather simple prompting approaches surpass extractive baselines and the majority of submitted CrisisFACTS systems.
This gives first insights into how openly available LLMs can be used for disaster summarization.
However, a qualitative analysis also reveals shortcomings and limitations of the proposed approach.
Interesting future directions include a detailed analysis of prompting strategies, the impact of query formulations, and how to address surface form issues.

\section*{Acknowledgments}

The authors acknowledge the financial support by the Federal Ministry of Education and Research of Germany in the project ISAKI (project number 13N15572).

\bibliography{anthology,custom}
\bibliographystyle{acl_natbib}

\appendix

\section{Human Evaluation}\label{appendix:metrics}
The submitted event nuggets for a system and event-request pair are ordered by importance and formed to a summary $\mathcal{S}$ by a rank cut-off $k$.
CrisisFACTS meta-facts are created by deduplicating all pooled participants event nuggets with $\text{BERTScore}(\cdot)$ F1-score.
Then, each meta-fact is annotated into Useful, Poor, Redundant, and Lagged categories.
Based on a  bipartite graph of event nuggets connected to the CrisisFACTS meta-facts, the comprehensiveness (i.e., recall) is calculated as

\begin{equation}
\frac{\sum \text{score of adjacent meta-facts}}{|\text{all meta-facts with non-zero score}|}
\end{equation}

Here the score of a meta-fact is predefined with $\text{Useful}=1.0$, $\text{Poor}=0.0$, $\text{Redundant}=0.5$, and $\text{Lagged}=0.0$.
Similarly, the redundancy (i.e., precision) is measured for a system and event-request pair as

\begin{equation}
\frac{\sum \text{score of adjacent meta-facts}}{|\text{all adjacent meta-facts}|}
\end{equation}

For the final results, all runs are macro-averaged across event-request pairs for an event, and then across all events.

\section{Prompting}\label{appendix:prompts}
We present the detailed prompt and response in \tablename~\ref{tab:app_sum_fact_extraction} and \tablename~\ref{tab:app_sum_fact_response}, respectively.

\begin{table*}[!htbp]
    \centering
    \begin{tabular}{p{0.95\linewidth}}
    \toprule
    You are a fact extractor for disaster response organizations. Use the documents to answer the question based on a list of extracted facts as evidence.
    \\\\
    Please follow the instructions for the facts:\\
    1. The facts must be short.\\
    2. The format of one fact is text-snippet (source document).\\
    3. Provide the source documents for each fact with the format: (Doc-1, Doc-2, ..)\\
    4. Include fact-relevant entities such as locations, numbers, dates, etc.\\
    5. Only include facts which are focused on the question.\\
    6. The list items must start with * bullet points. Do not use numberings.\\
    \\
    We provide you one example within `` marks: `\{demonstration\}`
    \\\\
    Your task
    \\
    Documents: \{documents\}
    \\
    Question: \{query\}
    \\\\
    Facts list:
    \\\bottomrule
    \end{tabular}
    \caption{Fact extraction}
    \label{tab:app_sum_fact_extraction}
\end{table*}

\begin{table*}[!htbp]
    \centering
    \begin{tabular}{p{0.95\linewidth}}
    \toprule
    Your task
    \\
    Documents: ...
    \\
    Question: How many firefighters are active?
    \\\\
    Facts list:
    \\
    * 3,300 firefighters are active in fighting the Lilac Fire (Doc-5)
    \\
    * 808 firefighters are battling the Skirball Fire in Los Angeles (Doc-7)
    \\
    * 6,946 firefighters are on scene fighting the Thomas Fire in Ventura and Santa Barbara (Doc-8)
    \\\\Answer: There are approximately 9,154 firefighters actively fighting fires in San Diego and neighboring counties.
    \\\bottomrule
    \end{tabular}
    \caption{Example response}
    \label{tab:app_sum_fact_response}
\end{table*}

\end{document}